\pdfoutput=1
\documentclass[11pt]{article}
\RequirePackage[OT1]{fontenc}
\RequirePackage{amsthm,amsmath}
\RequirePackage[numbers]{natbib}

\usepackage{indentfirst}
\usepackage[utf8]{inputenc} 
\usepackage[T1]{fontenc}    
\usepackage[colorlinks]{hyperref} 
\usepackage{url}            
\usepackage{booktabs}       
\usepackage{amsfonts}       
\usepackage{amsmath}
\usepackage{nicefrac}       
\usepackage{microtype}      
\usepackage{color}
\usepackage{bbm}
\usepackage{amssymb, enumerate}
\usepackage{makecell}
\usepackage{wrapfig}
\usepackage{extarrows}
\usepackage{arydshln}

\usepackage{caption}
\usepackage{multirow}

\usepackage{amsmath,amsfonts,amsthm, amssymb, enumerate}
\usepackage{lipsum}
\usepackage{caption}
\usepackage{amssymb,graphicx,subfigure}
\usepackage[bottom]{footmisc}
\usepackage[dvipsnames]{xcolor}
\usepackage{array,multirow}
\usepackage{enumitem}

\usepackage{microtype}
\usepackage{graphicx}
\usepackage{subfigure}
\usepackage{booktabs} 
\usepackage{amsmath, amssymb, amsfonts, amsthm} 
\usepackage{times}
\usepackage{epsfig}
\usepackage{footmisc}
\usepackage{enumitem}
\usepackage{setspace}
\usepackage{wrapfig}
\usepackage{extarrows}
\usepackage{makecell}
\usepackage{bbm}
\usepackage{textcomp}

\usepackage{caption} 
\usepackage{siunitx}
\usepackage{multirow} 

\usepackage{threeparttable} 
\usepackage{pifont}
\usepackage{hyperref}

\hypersetup{
    colorlinks=true,
    linkcolor=blue,
    filecolor=magenta,      
    urlcolor=cyan,
}

\usepackage{algorithm, algpseudocode}
\algnewcommand\TR{\item[{\textbf{Training phase}}]}
\algnewcommand\TE{\item[{\textbf{Test phase}}]}
\algnewcommand\Input{\item[{{Input:}}]}
\algnewcommand\Output{\item[{{Output:}}]}
\algnewcommand\Initialize{\item[{{Initialize:}}]}
\algnewcommand{\return}[1]{
	\State \textbf{return:}
	\Statex \hspace*{\algorithmicindent}\parbox[t]{.8\linewidth}{\raggedright #1}
}


\usepackage{setspace}

\begin{document}

\title{OFF-CLIP: Improving Normal Detection Confidence in Radiology CLIP with Simple Off-Diagonal Term Auto-Adjustment}
\date{}

\author{
    Junhyun Park$^{1}$\thanks{Equal contribution as first authors}, 
    Chanyu Moon$^{2}$\footnotemark[1], 
    Donghwan Lee$^{3}$, 
    Kyungsu Kim$^{3}$\thanks{Correspondence to Minho Hwang (\texttt{minho@dgist.ac.kr}) and Kyungsu Kim (\texttt{kyskim@snu.ac.kr})}, 
    Minho Hwang$^{1}$\footnotemark[2] \\
    \\
    {\footnotesize $^{1}$Department of Robotics and Mechatronics Engineering, DGIST, Republic of Korea} \\ 
    {\footnotesize \textit{\texttt{\{sean05071, minho\}@dgist.ac.kr}}} \\
    {\footnotesize $^{2}$Department of Electrical Engineering and Computer Science, DGIST, Republic of Korea} \\ 
    {\footnotesize \textit{\texttt{anscksdb0127@dgist.ac.kr}}}\\
    {\footnotesize $^{3}$Seoul National University, Republic of Korea} \\ 
    {\footnotesize \textit{\texttt{\{tdr.lee, kyskim\}@snu.ac.kr}}}
}

\maketitle
\begin{abstract}
Contrastive Language-Image Pre-Training (CLIP) has enabled zero-shot classification in radiology, reducing reliance on manual annotations. However, conventional contrastive learning struggles with normal case detection due to its strict intra-sample alignment, which disrupts normal sample clustering and leads to high false positives (FPs) and false negatives (FNs). To address these issues, we propose OFF-CLIP, a contrastive learning refinement that improves normal detection by introducing an off-diagonal term loss to enhance normal sample clustering and applying sentence-level text filtering to mitigate FNs by removing misaligned normal statements from abnormal reports. OFF-CLIP can be applied to radiology CLIP models without requiring any architectural modifications. Experimental results show that OFF-CLIP significantly improves normal classification, achieving a 0.61 Area under the curve (AUC) increase on VinDr-CXR over CARZero, the state-of-the-art zero-shot classification baseline, while maintaining or improving abnormal classification performance. Additionally, OFF-CLIP enhances zero-shot grounding by improving pointing game accuracy, confirming better anomaly localization. These results demonstrate OFF-CLIP’s effectiveness as a robust and efficient enhancement for medical vision-language models.


\end{abstract}

\section{Introduction}
Deep learning has significantly advanced medical imaging \cite{chan2020computer,jamshidi2020artificial}, but its dependence on large-scale annotated datasets limits its scalability. Zero-shot learning (ZSL) addresses this by enabling models to generalize without extensive manual labeling. Contrastive learning, particularly vision–language pretraining, has emerged as a strong paradigm for aligning large-scale image–text pairs \cite{clip_2021}. This approach has been adapted to radiology for zero-shot classification and anomaly detection \cite{gloria_2021,carzero_2024,chexzero_2022,medklip_2023,kad_2023,convirt_2022}.

ConVIRT \cite{convirt_2022} first explored aligning medical reports with images. GLoRIA \cite{gloria_2021} introduced text-weighted local attention, while KAD \cite{kad_2023} incorporated domain knowledge for entity-aware representations. CheXzero \cite{chexzero_2022} leveraged a CLIP-based framework for chest X-ray classification, and CARZero \cite{carzero_2024} improved abnormality alignment via cross-attention. 

\begin{figure}[t!]
\centering
\includegraphics[width=0.95\textwidth]{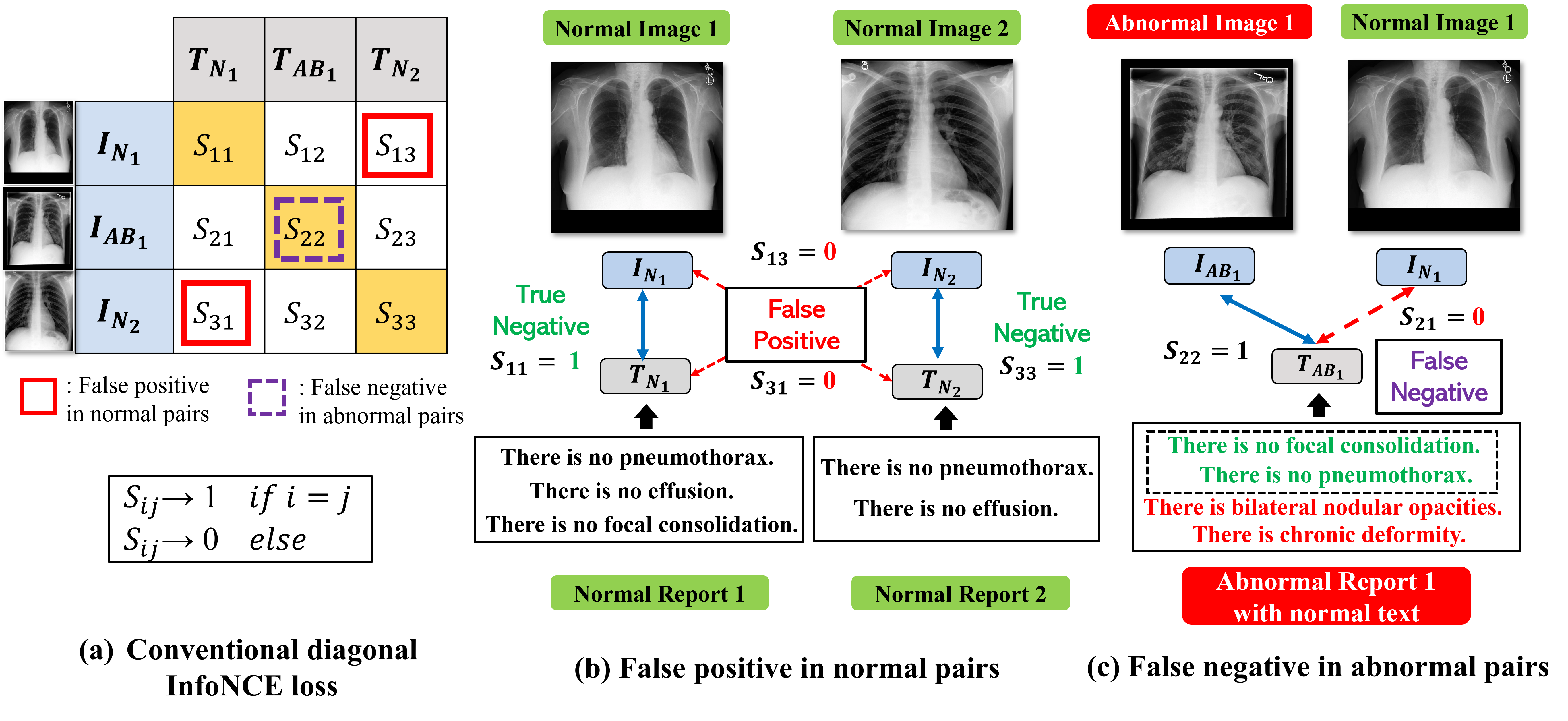}
\caption{The figure illustrates two key issues in (a) conventional diagonal InfoNCE loss in Radiology CLIP: (b) High false positives due to alignment restricted to matched pairs, forcing apart other normal samples, and (c) High false negatives caused by normal sentences in abnormal reports, bringing normal sentences closer to abnormal images while pushing them away from normal images.}
\label{fig1}
\end{figure}

These methods predominantly rely on InfoNCE loss, which aligns only matched image–text pairs in the $B \times B$ similarity matrix (where $B$ is the batch size). This strict diagonal assumption disregards relationships among normal samples, forcing semantically similar normal cases apart and leading to high false positives (FPs). (Fig. \ref{fig1}-(b)). Additionally, normal statements in abnormal reports introduce misalignment, increasing false negatives (FNs) (Fig. \ref{fig1}-(c)). As shown in Table~\ref{tab:comparison}, recent studies lack strategies to address these issues, often leading to normal AUCs below 0.5, making models unreliable for screening applications.

To address these challenges, we propose OFF-CLIP, a novel contrastive learning framework that refines medical image–text alignment. OFF-CLIP introduces an off-diagonal term loss to enhance normal clustering and a text filtering strategy to remove normal statements from abnormal reports, reducing both FPs and FNs. These modifications improve normal detection and anomaly localization, outperforming the CARZero baseline \cite{carzero_2024} in zero-shot classification.

\begin{itemize}
\item[\textbullet] We identify two key limitations in medical contrastive learning: (i) poor normal sample clustering leading to high FPs and (ii) misalignment from normal text in abnormal reports increasing FNs.
\item[\textbullet] We introduce (i) an off-diagonal term loss to enhance normal clustering and (ii) a text filtering strategy to mitigate misalignment, effectively reducing both FPs and FNs.
\item[\textbullet] We show that OFF-CLIP improves 0.5 on normal AUC while preserving abnormal AUC in zero-shot classification and enhances anomaly localization in the zero-shot grounding task, outperforming the CARZero baseline.
\end{itemize}

\begin{table}[t!]
\centering
\caption{Comparison of OFF-CLIP with Existing Radiology CLIP Models}
\label{tab:comparison}
\resizebox{0.95\linewidth}{!}{%
\begin{tabular}{ccccccc} 
\toprule
\textbf{Study} & \textbf{Text Input} & \begin{tabular}[c]{@{}c@{}} \textbf{Text} \\ \textbf{Encoder} \end{tabular} & \begin{tabular}[c]{@{}c@{}} \textbf{Img} \\  \textbf{Encoder} \end{tabular} & \begin{tabular}[c]{@{}c@{}} \textbf{CLIP} \\ \textbf{Loss} \end{tabular} & \textbf{Similarity} & \begin{tabular}[c]{@{}c@{}} \textbf{Require} \\ \textbf{Annotation?}  \end{tabular} \\ 
\midrule
ConVIRT \cite{convirt_2022} & \begin{tabular}[c]{@{}c@{}} Rand. sent. \\ select \end{tabular} & \begin{tabular}[c]{@{}c@{}} Clinical \\ BERT \cite{clinicalbert} \end{tabular} & ResNet-50 & InfoNCE & Cosine & No \\ 
\hline
CheXZero \cite{chexzero_2022} & \begin{tabular}[c]{@{}c@{}} Impression \\ sec. \end{tabular} & \begin{tabular}[c]{@{}c@{}} Transformer \\ (CLIP) \cite{clip_2021} \end{tabular} & ViT-B/32 & InfoNCE & Cosine & No \\ 
\hline
GLoRIA \cite{gloria_2021} & Full report & \begin{tabular}[c]{@{}c@{}} BioClinical \\ BERT \cite{bioclinicalbert_2019} \end{tabular} & ResNet-50 & InfoNCE & Cosine & No \\ 
\hline
KAD \cite{kad_2023} & \begin{tabular}[c]{@{}c@{}} Extracted \\ entities \end{tabular} & \begin{tabular}[c]{@{}c@{}} PubMed \\ BERT \cite{pubmedbert_2021} \end{tabular} & \begin{tabular}[c]{@{}c@{}} ResNet-50, \\ ViT-16 \end{tabular} & \begin{tabular}[c]{@{}c@{}} InfoNCE, \\ BCE \end{tabular} & Cosine & No \\ 
\hline
CARZero \cite{carzero_2024} & \begin{tabular}[c]{@{}c@{}} Prompting \\  + rand. select \end{tabular}& \begin{tabular}[c]{@{}c@{}} Bio \\ BERT \cite{lee2020biobert} \end{tabular} & ViT-B/16 & InfoNCE & Cross-Attn & No \\ 
\midrule
\begin{tabular}[c]{@{}c@{}} \textbf{OFF-CLIP} \\ \textbf{(Ours)}  \end{tabular} & \begin{tabular}[c]{@{}c@{}} Prompting \\ + rand. select \\ + \textbf{Text filter} \end{tabular} & \begin{tabular}[c]{@{}c@{}} Bio \\ BERT \cite{lee2020biobert} \end{tabular} & \begin{tabular}[c]{@{}c@{}} ViT-B/16 \end{tabular} & \begin{tabular}[c]{@{}c@{}} \textbf{Off-diag.} \\ \textbf{\& abn.} \\ \textbf{InfoNCE} \end{tabular} & \begin{tabular}[c]{@{}c@{}} Agnostic \\ (Cosine \& \\ Cross-Attn) \end{tabular} & \begin{tabular}[c]{@{}c@{}} \textbf{No} \\ (pseudo- \\ label) \end{tabular}\\
\bottomrule
\end{tabular}
}
\end{table}

\section{Method}
\subsection{Image and Text Encodings}
We employ a text encoder and an image encoder to extract global and local feature representations. OFF-CLIP supports both ViT-based \cite{vit_2021} and ResNet-based \cite{resnet_2016} models. In this study, we use ViT-B/16 pretrained with M3AE \cite{M3AE_2022}, following the CARZero baseline. For text encoding, any pretrained BERT model can be used, and we adopt BioBERT \cite{lee2020biobert} for its domain-specific advantages.

\subsection{Off-Diagonal Term Loss and Abnormal InfoNCE Loss}
\subsubsection{Baseline Loss}
The conventional InfoNCE loss is formulated as:
\begin{equation}
    \mathcal{L}_{\text{baseline}} = - \frac{1}{B} \sum_{i=1}^{B} \left( \log \frac{e^{S^{i,i}}}{\sum_{j=1}^{B} e^{S^{i,j}}} + \log \frac{e^{S^{i,i}}}{\sum_{j=1}^{B} e^{S^{j,i}}} \right).
\label{eq:baseline}
\end{equation}
where $B$ is the batch size, and $S \in \mathbb{R}^{B \times B}$ is the similarity matrix between global and local image-text feature vectors. Different methods can be used to construct $S$, and we adopt a cross-attention-based approach \cite{carzero_2024} in this study.

\begin{figure}[t!]
\centering
\includegraphics[width=0.95\textwidth]{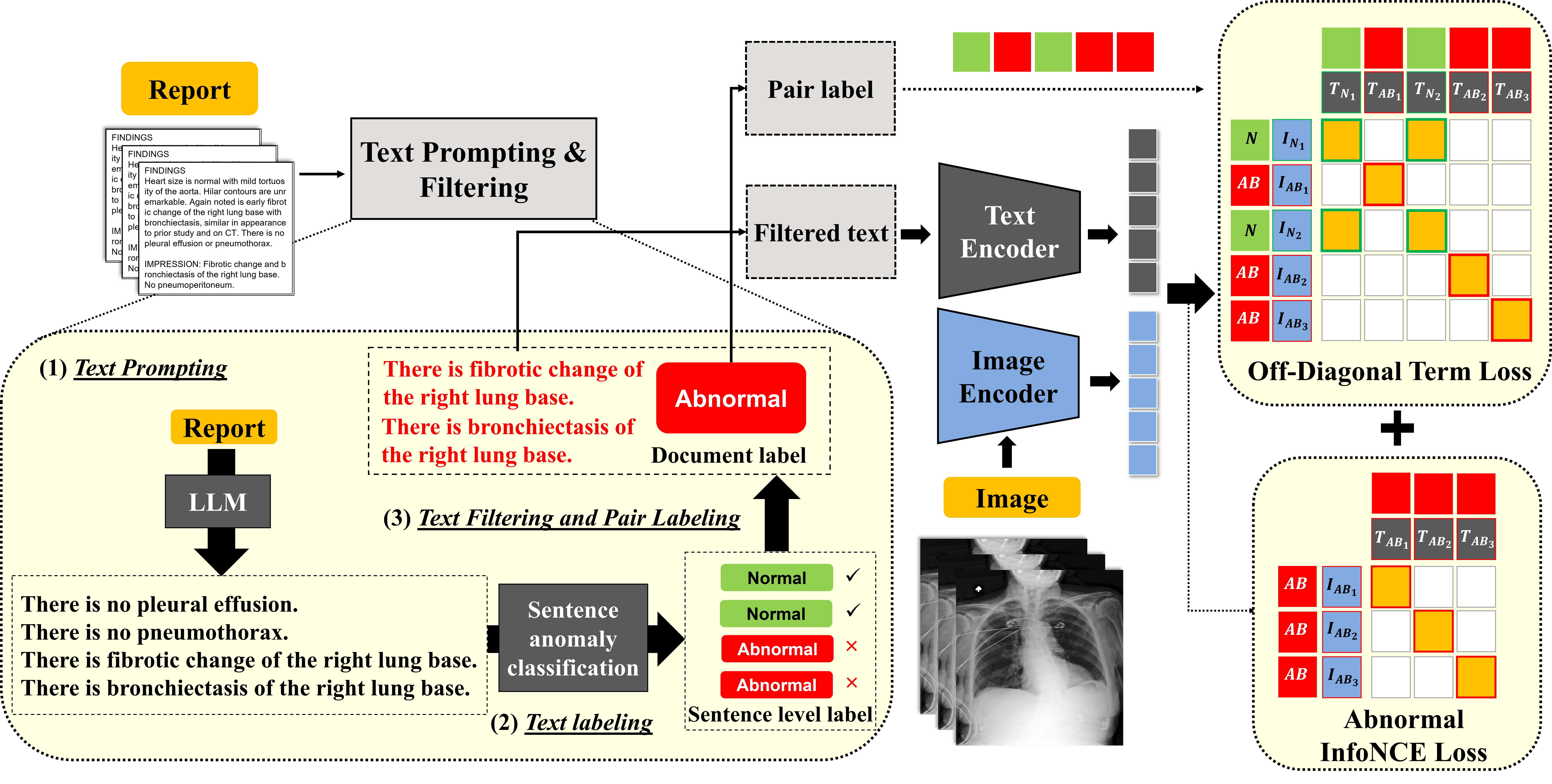}
\caption{OFF-CLIP leverages an off-diagonal term loss to effectively cluster normal samples within a batch. Abnormal pairs are further refined using an abnormal-only InfoNCE loss. Reports are processed using an LLM for text prompting, and sentence-level anomaly classification is applied to label each sentence. Normal sentences in abnormal reports are then filtered to reduce misalignment.}

\label{fig2}
\end{figure}

\subsubsection{Off-Diagonal Term Loss}
\label{sec:off_digonal}
The baseline loss \eqref{eq:baseline} relies on intra-sample image-text associations, misaligning normal representations and increasing false positives (FPs). To mitigate this, we introduce an off-diagonal term loss that clusters normal samples within a batch, formulated as:
\begin{align}
\mathcal{L}_{\text{off}} = & - \frac{1}{2B^2} \sum_{i=1}^{B} \sum_{j=1}^{B} 
\Big( \hat{Y}_{i,j} \log \sigma(S_{i,j}) + (1 - \hat{Y}_{i,j}) \log (1 - \sigma(S_{i,j})) \Big) \nonumber \\
& + \Big( \hat{Y}_{j,i} \log \sigma(S_{j,i}) + (1 - \hat{Y}_{j,i}) \log (1 - \sigma(S_{j,i})) \Big).
\end{align}

where $S \in \mathbb{R}^{B \times B}$ is the similarity matrix between image and text latents, $\sigma(S_{i,j}) = \frac{1}{1 + e^{-S_{i,j}}}$ is the sigmoid activation, and $\hat{Y}_{i,j}$ is defined as:  
\begin{equation}
\hat{Y}_{ij} =
\begin{cases} 
1, & \text{if } i = j \text{ (diagonal) or both } i, j \text{ are pseudo-normal pairs ($\hat{n}$)}    \\ 
0, & \text{otherwise}.
\end{cases}
\end{equation}

Pseudo-labels for each pair ($\hat{n}$, $\hat{a}$) are generated using a pretrained radiology sentence-level anomaly classifier \cite{S_KD_2024}, as detailed in Section \ref{sec:text_prompting}, without human annotation. This formulation extends CLIP loss by reinforcing positive alignment among normal pairs while preserving standard diagonal alignment for all pairs, including abnormal ones.

\subsubsection{Abnormal InfoNCE Loss}
The off-diagonal term loss increases the number of normal labels, which may reduce sensitivity to abnormal cases. To counteract this, we apply the original InfoNCE loss to abnormal pairs only. Let $A$ be the number of abnormal pairs in the batch. The abnormal similarity matrix $S_{ab} \in \mathbb{R}^{A \times A}$ is extracted from the full similarity matrix $S$, containing only abnormal similarity scores:
\begin{equation}
    \mathcal{L}_{\text{ab}} = - \frac{1}{A} \sum_{i=1}^{A} \left( \log \frac{e^{S_{ab}^{i,i}}}{\sum_{j=1}^{A} e^{S_{ab}^{i,j}}} + \log \frac{e^{S_{ab}^{i,i}}}{\sum_{j=1}^{A} e^{S_{ab}^{j,i}}} \right).
\end{equation}

\subsubsection{Total Loss}
The total loss of OFF-CLIP is defined as:
\begin{equation}
    \mathcal{L}_{\text{OFF-CLIP}} = \mathcal{L}_{\text{off}} + \lambda_{\text{ab}} \mathcal{L}_{\text{ab}}
\label{eq:off_clip}
\end{equation}

where $\lambda_{\text{ab}}$ (set to 1) balances the abnormal InfoNCE loss, preserving normal sample clustering while maintaining sensitivity to abnormal cases.

\subsection{Text Prompting and Filtering}
\label{sec:text_prompting}
\subsubsection{Text Prompting}
We employ the GPT-4o model to extract the Findings and Impressions sections from medical reports. The extracted text is reformatted into a structured template: There is {disease} (e.g., There is no pneumonia, There is pleural effusion at the left lung base). This structured format follows CARZero \cite{carzero_2024}, which has demonstrated state-of-the-art performance in zero-shot classification.

\begin{table}[t!]
\centering
\caption{Zero-shot classification performance evaluated using AUC. OFF-CLIP integrates text filtering, off-diagonal term loss, and abnormal InfoNCE loss. We evaluate it on VinDr-CXR, Open-I, PadChest, and CheXpert, applying OFF-CLIP to CARZero.  Note that results may differ from the original baseline paper as we retrained CARZero to ensure consistency.}
\label{tab:baseline_test}
\begin{threeparttable}
\begin{tabular}{ccccccccccc} 
\toprule
Dataset                     & \begin{tabular}[c]{@{}c@{}}OFF\\-CLIP\end{tabular} & Normal        & ATE           & CM            & EMPH          & EFF           & NOD           & PLT           & PNA           & TOT            \\ 
\hline
\multirow{2}{*}{VinDr-CXR~} & \ding{55}                         & 0.25          & 0.74          & 0.91          & 0.93          & 0.94          & 0.47          & 0.75          & \textbf{0.90} & 0.79           \\
                            & \ding{51}                         & \textbf{0.86} & \textbf{0.84} & \textbf{0.93} & \textbf{0.95} & \textbf{0.94} & \textbf{0.86} & \textbf{0.86} & 0.83          & \textbf{0.87}  \\ 
\hline
\multirow{2}{*}{Open-I}     & \ding{55}                        & 0.32          & 0.79          & 0.89          & 0.83          & 0.93          & 0.55          & 0.73          & 0.84          & 0.72           \\
                            & \ding{51}                        & \textbf{0.74} & \textbf{0.81} & \textbf{0.90} & \textbf{0.91} & \textbf{0.93} & \textbf{0.60} & \textbf{0.76} & \textbf{0.84} & \textbf{0.81}  \\ 
\hline
\multirow{2}{*}{PadChest}   & \ding{55}                        & 0.24          & 0.78          & 0.87          & 0.78          & 0.95          & 0.52          & 0.66          & 0.80          & 0.68           \\
                            & \ding{51}                         & \textbf{0.77} & \textbf{0.82} & \textbf{0.88} & \textbf{0.89} & \textbf{0.95} & \textbf{0.64} & \textbf{0.81} & \textbf{0.80} & \textbf{0.76}  \\ 
\hline
\multirow{2}{*}{CheXpert}   & \ding{55}                         & 0.38          & 0.73          & 0.86          & -             & 0.92          & -             & -             & 0.65          & 0.73           \\
                            & \ding{51}                         & \textbf{0.81} & \textbf{0.87} & \textbf{0.88} & -             & \textbf{0.92} & -             & -             & \textbf{0.79} & \textbf{0.86}  \\
\bottomrule
\end{tabular}
\begin{tablenotes}
    \item[*] ATE, CM, EMPH, EFF, NOD, PLT, PNA, and TOT represent Atelectasis, Cardiomegaly, Emphysema, Effusion, Nodule, Pleural Thickening, Pneumonia, and Total, respectively. 
\end{tablenotes}
\end{threeparttable}
\end{table}

\subsubsection{Pseudo-Label Extraction}
Since CLIP training operates in a zero-supervision setting, direct human supervision is not feasible. To address this, we use a pretrained sentence-level anomaly detection model for radiology \cite{S_KD_2024} to assign pseudo-labels at both the sentence and report levels. Each sentence is labeled as normal, abnormal, or uncertain, and the report-level pseudo-label is determined by aggregating these sentence labels. A report is classified as abnormal ($\hat{a}_{report}$) if it contains at least one abnormal sentence, whereas it is labeled as normal ($\hat{n}_{report}$) if all sentences are either normal or uncertain. The report-level pseudo-label ($\hat{n}_{report}$, $\hat{a}_{report}$) is then assigned to each image-text pair ($\hat{a}$, $\hat{n}$), ensuring that the image and its corresponding report share the same label for contrastive learning.

\subsubsection{Text Filtering for False Negative Reduction} To reduce false negatives, we eliminate normal and uncertain sentences from abnormal reports. This prevents contrastive learning from misaligning normal descriptions with abnormal images or pushing normal descriptions away from normal images. By ensuring accurate abnormal image-text associations, this filtering step enhances anomaly detection.

\begin{table}[t!]
\centering
\caption{Ablation Study on CARZero Baseline for Zero-Shot Classification Performance using the Open-I Dataset, and evaluated by AUC. By training the baseline with 6 different options, and we validated on the Open-I dataset.}
\label{tab:open_I_carzero_AUC}
\resizebox{0.95\linewidth}{!}{%
\begin{threeparttable}

\begin{tabular}{ccccccccccccc} 
\toprule
\begin{tabular}[c]{@{}c@{}}Text \\Filter\end{tabular} & $\mathcal{L}_{off}$ & \multicolumn{1}{l}{$\mathcal{L}_{ab}$} & Normal        & ATE           & CALC          & CM           & EMPH          & MASS          & NOD           & OPAC          & PLT           & Total          \\ 
\hline
               &                  &                                        & 0.32          & 0.79          & 0.59          & 0.89          & 0.83          & 0.68          & 0.55          & \textbf{0.79} & 0.73          & 0.72           \\
 & {\ding{51}}                 &                                        & 0.72          & 0.29          & 0.49          & 0.34          & 0.33          & 0.31          & 0.47          & 0.24          & 0.28          & 0.32           \\
       & {\ding{51}}                 & {\ding{51}}                                       & \textbf{0.75} & 0.24          & 0.46          & 0.30          & 0.27          & 0.27          & 0.44          & 0.23          & 0.26          & 0.29           \\
{\ding{51}}                                                     &                  &                                        & 0.62          & 0.58          & 0.50          & 0.64          & 0.46          & 0.27          & 0.49          & 0.31          & 0.41          & 0.44           \\
{\ding{51}}                                                     & {\ding{51}}                 &                                        & 0.72          & 0.77          & 0.55          & 0.84          & 0.85          & 0.76          & \textbf{0.61} & 0.75          & 0.74          & 0.78           \\
{\ding{51}}                                                     & {\ding{51}}                 & {\ding{51}}                                       & 0.74          & \textbf{0.81} & \textbf{0.60} & \textbf{0.90} & \textbf{0.91} & \textbf{0.85} & 0.60          & 0.73          & \textbf{0.76} & \textbf{0.81}  \\
\bottomrule
\end{tabular}
\begin{tablenotes}
    \item[*] CALC, MASS, OPAC refer to Calcification, Mass and Opacity, respectively. And the another abbreviation are defined on previous table.
\end{tablenotes}
\end{threeparttable}
}
\end{table}

\begin{table}[t!]
\centering
\caption{Ablation study on the CARZero baseline for zero-shot classification performance on the Open-I dataset. We report false positive (FP) and false negative (FN) rates, along with their proportions relative to the total sample count. Additionally, we assess the FP-FN balance and demonstrate how text filtering mitigates FN errors. FN reduction is quantified as the relative improvement when text filtering is enabled.}
\label{tab:open_I_carzero_fn}
\resizebox{0.95\linewidth}{!}{%
\begin{tabular}{c ccccccccc} 
\toprule
row\# & \begin{tabular}[c]{@{}c@{}}Text \\Filter\end{tabular} & $\mathcal{L}_{off}$ & $\mathcal{L}_{ab}$ & $\frac{\text{FN}}{\text{Total}}$ & $\frac{\text{FP}}{\text{Total}}$ & $\frac{\text{FN}}{\text{FP+FN}}$ & $\frac{\text{FP}}{\text{FP+FN}}$ & |$\frac{\text{FN}}{\text{FP+FN}}$ - $\frac{\text{FP}}{\text{FP+FN}}$| & $\frac{\text{FN}}{\text{Total}}$ reduction \\ 
\hline
1 & &  &  & 0.0003  & 0.37  & 0.0007  & \textbf{0.9993}  & \textbf{0.9986} & N/A (abn. biased) \\ 
 2 & & {\ding{51}} &  & 0.14  & 0.16  & 0.47  & 0.53  & 0.06  & - \\ 
3 & & {\ding{51}} & {\ding{51}} & 0.17  & 0.13  & 0.57  & 0.43  & 0.14  & - \\  
4 & {\ding{51}} &  &  & 0.08  & 0.22  & 0.27  & 0.73  & 0.46  & - \\  
5 & {\ding{51}} & {\ding{51}} &  & 0.13  & 0.17  & 0.43  & 0.57  & 0.14  & \textbf{↓7.1\%} (vs. row 2) \\  
6 & {\ding{51}} & {\ding{51}} & {\ding{51}} & 0.15  & 0.14  & \textbf{0.51}  & \textbf{0.50}  & \textbf{0.01}  & \textbf{↓11.8\%} (vs. row 3) \\ 
\bottomrule
\end{tabular}
}
\end{table}

\section{Experimental Results}
\subsection{Datasets}
We train OFF-CLIP using the MIMIC-CXR dataset \cite{mimic_cxr}, which contains 377,110 chest radiographs with associated reports. Only frontal images are retained; for studies with multiple frontal images, one is randomly selected, and for reports with several prompted sentences, one is chosen per epoch. Training is further restricted to images from the p10-16 folders within the p10-19 range.

For evaluation and ablation, we use four datasets: VinDr-CXR \cite{vindr_cxr} (18,000 X-rays, 28 disease annotations with bounding boxes; 3,000 evaluation scans, 68.3\% normal), Open-I \cite{open_I} (7,470 X-rays, 18 disease annotations), CheXpert \cite{chexpert_2019} (224,316 X-rays, 14 disease annotations; evaluated on 500 cases), and PadChest \cite{padchest_2020} (160,868 X-rays, 192 disease annotations; evaluated on 39,053 manually labeled cases). All datasets are publicly available. MIMIC-CXR, VinDr-CXR, and CheXpert require access approval from PhysioNet.

\subsection{Zero-Shot Classification Performance}
Table \ref{tab:baseline_test} compares the zero-shot classification performance of the state-of-the-art CARZero baseline with and without OFF-CLIP across four test sets, using AUC as the evaluation metric. OFF-CLIP achieves a 177\% relative gain in average normal AUC over CARZero, improving it by 0.50. Additionally, AUC scores for multiple chest disease categories (Atelectasis, Cardiomegaly, Emphysema, Effusion, Nodule, Pleural Thickening) show notable gains, leading to a total AUC improvement of 0.095. These results confirm that OFF-CLIP effectively reduces false negatives caused by strict loss constraints and false positives from normal sentences in abnormal reports.

\begin{table}[t!]
\centering
\caption{Zero-shot grounding performance comparison using the pointing game with top-10\% and top-20\% attention on the VinDr-CXR dataset, evaluated for the CARZero baseline and CARZero with OFF-CLIP. Attention maps were assessed for each disease-specific prompt.}
\label{tab:pointing_game}
\resizebox{\linewidth}{!}{%
\begin{threeparttable}
\begin{tabular}{lcccccccccccc} 
\toprule
Attention                  & Methods  & AE           & ATE           & CALC          & CM          & CONS          & EFF           & INF           & NOD           & OPAC          & PLT            & PTX           \\ 
\hline
\multirow{2}{*}{High 10\%} & Baseline & 0.94         & 0.61          & 0.30          & 0.99         & 0.84          & 0.69          & 0.8           & 0.47          & 0.66          & 0.33          & 0.63          \\
                           & OFF-CLIP & \textbf{1.0} & \textbf{0.78} & \textbf{0.59} & \textbf{1.0} & \textbf{0.95} & \textbf{0.89} & \textbf{0.9}  & \textbf{0.73} & \textbf{0.92} & \textbf{0.47} & \textbf{1.0}  \\ 
\hline
\multirow{2}{*}{High 20\%} & Baseline & 0.99         & 0.78          & 0.59          & 1.0          & 0.97          & 0.79          & 0.9           & 0.64          & 0.83          & 0.47          & 0.63          \\
                           & OFF-CLIP & \textbf{1.0} & \textbf{0.89} & \textbf{0.82} & \textbf{1.0} & \textbf{0.97} & \textbf{1.0}  & \textbf{0.95} & \textbf{0.84} & \textbf{0.96} & \textbf{0.67} & \textbf{1.0}  \\
\bottomrule
\end{tabular}
\begin{tablenotes}
    \item[*] AE, CONS, and INF denote Aortic Enlargement, Consolidation, and Infiltration, respectively. Other abbreviations are defined in the previous table.
\end{tablenotes}
\end{threeparttable}
}
\end{table}

\begin{figure}[t!]
\centering
\includegraphics[width=0.85\textwidth]{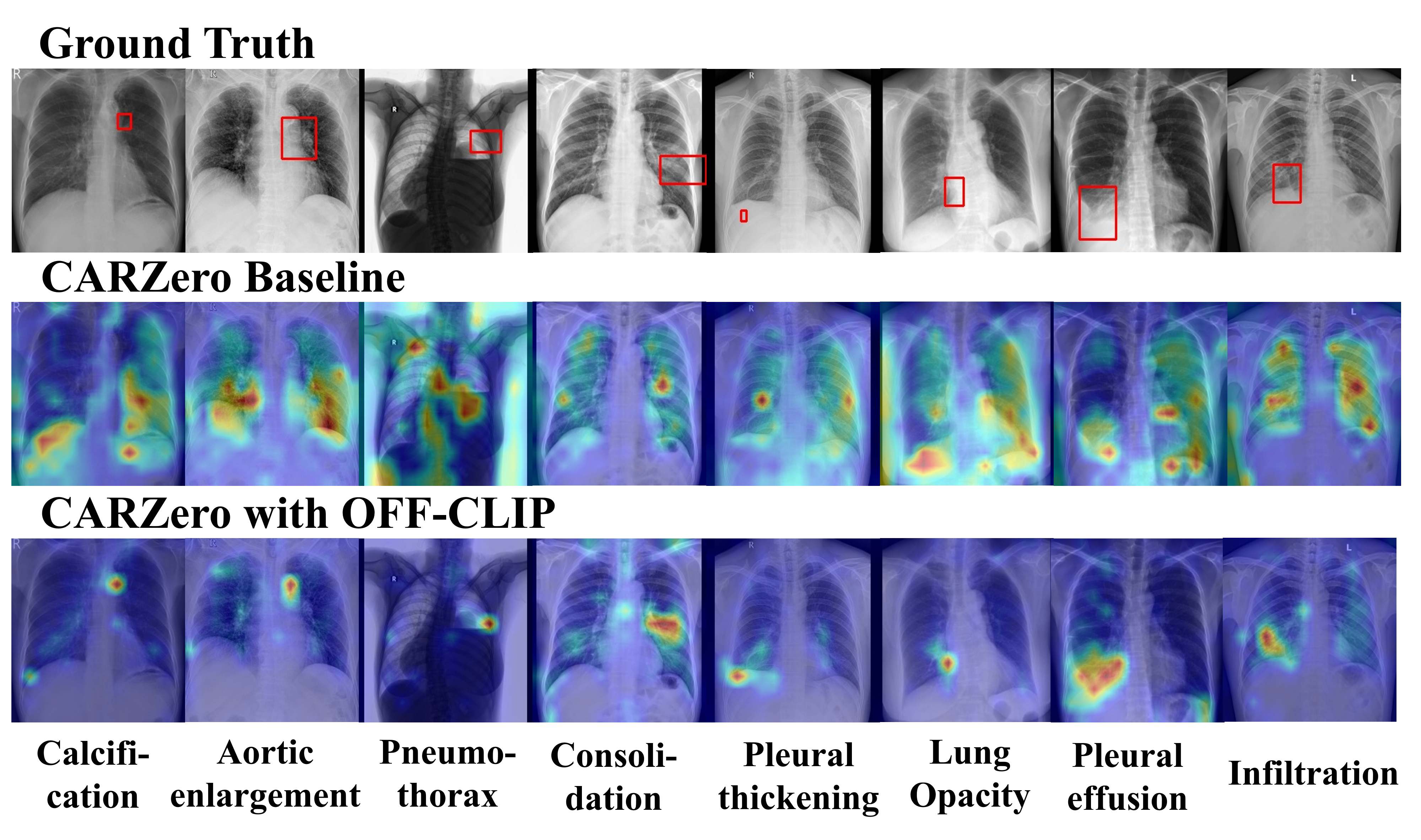}
\caption{Visualization of attention maps on VinDr-CXR. Red boxes indicate ground truth bounding boxes for each diseases. Highlighted pixels represent regions with higher activation weights, linking specific words to image areas.} 
\label{fig3}
\end{figure}

\subsection{Ablation Study on Zero-shot classification}
We conduct an ablation study on Open-I with CARZero to assess the contributions of OFF-CLIP's components: text filtering, the off-diagonal loss (\(\mathcal{L}_{off}\)), and the abnormal pair InfoNCE loss (\(\mathcal{L}_{ab}\)). 

Table \ref{tab:open_I_carzero_AUC} shows that \(\mathcal{L}_{off}\) significantly enhances normal detection, maintaining AUC above 0.7 compared to 0.3 in the baseline. Text filtering further mitigates FN errors, particularly when combined with \(\mathcal{L}_{off}\) and \(\mathcal{L}_{ab}\).  Table \ref{tab:open_I_carzero_fn} presents FN and FP ratios and rates. Without OFF-CLIP, the model is heavily biased toward abnormal cases, with FP rates nearing 1. Introducing \(\mathcal{L}_{off}\) balances FP and FN misclassifications, reducing FP rates from 0.99 to 0.5. Text filtering further lowers FN ratios (\(\mathcal{L}_{off} + \mathcal{L}_{ab}\): 0.17 → 0.15) while maintaining a balanced FP-FN distribution. These findings confirm that conventional contrastive learning struggles with normal clustering and misalignment, while OFF-CLIP effectively mitigates these issues, improving zero-shot classification reliability.  

\vspace{-2mm}
\subsection{Zero-shot Grounding Performance}  
We evaluate anomaly localization using the pointing game \cite{pointing_game}, selecting the top 10\% or 20\% high-attention regions instead of only the highest one. A sample is considered successful if any selected region overlaps with the ground truth bounding box, and the average success rate is computed per disease.

Table \ref{tab:pointing_game} demonstrates that OFF-CLIP significantly improves attention alignment over the CARZero baseline. With 10\% selection, CALC improves from 0.30 to 0.59 (+97\%), while with 20\%, PTX increases from 0.63 to 1.00 (+58.73\%). Figure \ref{fig3} further reveals that while CARZero’s attention is diffuse, OFF-CLIP’s is sharply focused on ground truth regions. These results confirm that OFF-CLIP enhances zero-shot grounding and anomaly localization.

\vspace{-3mm}
\section{Conclusion}
OFF-CLIP addresses two key limitations in radiology CLIP models: excessive false positives from poor normal sample clustering and high false negatives due to misleading normal text in abnormal reports. Through extensive ablation studies, we quantified their impact using false positive and false negative analyses.

By incorporating an off-diagonal similarity loss (\(\mathcal{L}_{off}\)) and a text filtering strategy, OFF-CLIP significantly improves normal detection (0.50 AUC, 177\% average increase) and enhances anomaly localization (up to 97\% improvement in attention alignment). While OFF-CLIP is evaluated on CARZero, its framework-agnostic design suggests potential applicability to other models. Further validation is required to assess its effectiveness across different architectures and its clinical utility in real-world settings.

\section*{Acknowledgments}
Kyungsu Kim is affiliated with the School of Transdisciplinary Innovations, Department of Biomedical Science, Medical Research Center, Interdisciplinary Program in Bioengineering, and Interdisciplinary Program in Artificial Intelligence at Seoul National University, Seoul, Republic of Korea. Donghwan Lee is affiliated with Department of Biomedical Science at Seoul National University, Seoul, Republic of Korea.

This work was supported by a grant from the Institute of Information \& Communications Technology Planning \& Evaluation (IITP), funded by the Korean government (MSIT) (RS2021-II211343), Artificial Intelligence Graduate School Program, Seoul National University. This work was supported by Korea Medical Device Development Fund grant funded by the Korea government (1711196477, RS‐2023‐00252244), the National Research Council of Science \& Technology (NST) grant funded by the Korea government (MSIT) (CRC23021‐000), the Industrial Strategic Technology Development Program (ISTDP) (RS-2024-00443054) funded by the Ministry of Trade, Industry \& Energy (MOTIE, Korea), and by the collaborative project with ROEN Surgical Inc.

\bibliographystyle{splncs04}
\bibliography{citation}

\end{document}